\documentclass[11pt]{article}

\usepackage[preprint]{acl}

\usepackage{times}
\usepackage{latexsym}
\usepackage[T1]{fontenc}
\usepackage[utf8]{inputenc}
\usepackage{microtype}
\usepackage{inconsolata}
\usepackage{graphicx}
\usepackage{booktabs}
\usepackage{array}
\usepackage{amsmath}
\usepackage{amssymb}
\usepackage{tikz}
\usetikzlibrary{positioning,arrows.meta,fit,calc}
\usepackage{pgfplots}
\pgfplotsset{compat=1.18}
\usepackage{colortbl}
\usepackage{capt-of}
\definecolor{moss}{HTML}{454D42}
\definecolor{mosssoft}{HTML}{E5EADE}
\definecolor{plotgreen}{HTML}{3F7D52}
\definecolor{gold}{HTML}{C28230}
\definecolor{inset}{HTML}{E7E2DC}
\definecolor{basegrey}{HTML}{9CA3AF}

\tikzset{
  hf-process/.style={draw=moss, line width=0.7pt, rounded corners=2.5pt,
    fill=white, align=center, font=\small, inner sep=5pt, minimum height=0.92cm},
  hf-artifact/.style={draw=moss!55, line width=0.65pt, rounded corners=2.5pt,
    fill=mosssoft, align=center, font=\small, inner sep=5pt, minimum height=0.76cm},
  hf-flow/.style={-{Latex[length=1.8mm]}, draw=moss, line width=0.8pt},
  hf-feedback/.style={-{Latex[length=1.8mm]}, draw=gold, dashed, line width=0.9pt},
  hf-note/.style={font=\scriptsize, text=black!60, align=center},
  hf-stage/.style={font=\scriptsize\bfseries, text=moss}
}

\title{Microfitting: Continual Learning via Cached Test-Time Training}

\author{
  Nick Ryan, Peter A. Devine, Ben Sirb, Hannah Sansford, Alex Chiocchi \\
  Interval \\
  \texttt{\{nick.ryan, peter.devine, ben, hannah, alex\}@ai-interval.com}
}

\begin{document}
\maketitle
\begin{abstract}
Continual learning is a promising approach to specialize language models to an organization's private data.
However, current methods exhibit a number of drawbacks including catastrophic forgetting, where new training
updates overwrite prior knowledge. We present microfitting, which trains and caches a small adapter for each
document. Each adapter is trained once, offline, on synthetic questions targeted at demonstrated base-model
shortcomings and reused whenever that document is queried. This method borrows from test-time learning literature
and sidesteps the problem of continual learning through parameter isolation, allowing the system to specialize on
new data without overwriting prior knowledge. The process that generates adapter training data also improves over
time, revising its recipe from previous outcomes and fresh examples of the target workload. On DeepSeek V4 Flash,
microfitting substantially improves accuracy over the base model. In Enterprise Document Operations (EDO), an
open-book Q\&A environment over real-world enterprise documents, our method achieves a 38.2\% relative improvement
over base-model performance. Microfitting makes
continual specialization practical and cost-efficient:
a frozen model can adapt to an evolving private corpus overnight on local hardware, without modifying the base
model or previously cached adapters.

\end{abstract}

\begin{figure}[t]
\centering
\begin{tikzpicture}
\begin{axis}[
  width=\columnwidth, height=5.8cm,
  xmin=-0.15, xmax=3.15, xtick={0,1,2,3},
  ymin=0, ymax=50, ytick={0,10,20,30,40,50},
  ylabel={Improvement over baseline (\%)},
  xlabel={Pipeline iteration},
  axis lines=left, tick align=outside,
  tick label style={font=\footnotesize}, label style={font=\footnotesize},
  ymajorgrids, grid style={black!12}, axis line style={black!55}
]
\addplot[color=plotgreen, line width=1.1pt, mark=*, mark size=2.4pt,
  mark options={solid, fill=plotgreen}]
  coordinates {(0,0) (1,30.9) (2,37.5) (3,38.2)};
\node[font=\scriptsize, anchor=south, yshift=5pt] at (axis cs:1,30.9) {30.9\%};
\node[font=\scriptsize, anchor=south, yshift=5pt] at (axis cs:2,37.5) {37.5\%};
\node[font=\scriptsize, anchor=south east, xshift=3pt, yshift=5pt] at (axis cs:3,38.2) {38.2\%};
\end{axis}
\end{tikzpicture}
\caption{Automated refinement of the microfitting pipeline. The system enters a novel data environment and teaches
itself to produce training data for new per-document adapters designed to outperform the base model. The system
automatically improves the data generation process in EDO through
iterative refinement based on previous adapter performance.}
\label{fig:headline}
\end{figure}
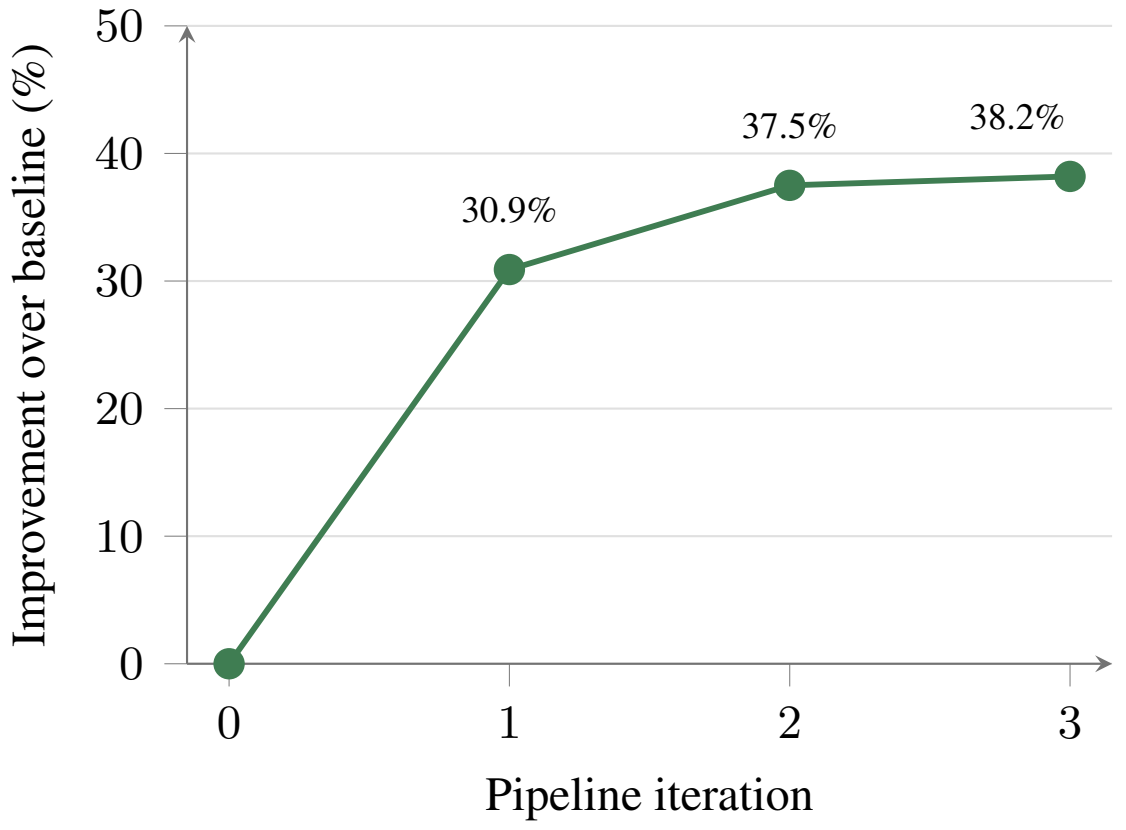

\begin{figure*}[t]
\centering
\begin{tikzpicture}[
  x=1cm, y=1cm,
  flow/.style={hf-flow},
  newflow/.style={hf-flow, draw=gold},
  box/.style={hf-process, inner sep=6pt},
  adapter/.style={hf-artifact, minimum width=2.05cm,
    minimum height=0.43cm, inner sep=2pt},
  stage/.style={hf-stage, font=\small\bfseries},
  note/.style={hf-note}
]
\node[stage] at (4.65,5.05) {TRAIN ONCE ON LOCAL HARDWARE};
\node[stage] at (12.6,5.05) {REUSE ON EVERY QUERY};

\draw[draw=gold, fill=gold!8, line width=0.8pt]
  (1.25,3.05) -- (1.25,4.12) -- (1.96,4.12) -- (2.20,3.88)
  -- (2.20,3.05) -- cycle;
\draw[draw=gold, line width=0.65pt] (1.96,4.12) -- (1.96,3.88) -- (2.20,3.88);
\foreach \y in {3.65,3.47,3.29} {
  \draw[gold!65, line width=0.6pt] (1.43,\y) -- (2.01,\y);
}
\node[font=\small, text=gold!80!black] at (1.72,2.76) {New document};
\node[box, minimum width=3.15cm, minimum height=0.98cm]
  (training) at (1.72,1.42) {Offline edge training\\[2pt]
    {\scriptsize One epoch during idle hours}};
\draw[newflow] (1.72,2.53) -- (training.north);

\draw[draw=moss!35, fill=black!1, rounded corners=4pt, line width=0.65pt]
  (4.25,0.73) rectangle (9.25,4.52);
\node[font=\small\bfseries, text=moss] at (6.75,4.20) {Growing adapter library};
\foreach \idx/\y in {1/3.53,2/2.82,3/2.11} {
  \draw[draw=moss!50, fill=white, line width=0.6pt]
    (4.65,\y-0.22) rectangle (4.98,\y+0.22);
  \draw[moss!45, line width=0.45pt]
    (4.72,\y+0.08) -- (4.91,\y+0.08)
    (4.72,\y-0.02) -- (4.91,\y-0.02)
    (4.72,\y-0.12) -- (4.87,\y-0.12);
  \node[font=\small, anchor=west] at (5.12,\y) {Document \idx};
  \node[adapter] at (8.02,\y) {Adapter \idx};
}
\draw[draw=gold!55, fill=gold!10, rounded corners=2pt, line width=0.7pt]
  (4.43,1.05) rectangle (9.07,1.79);
\draw[draw=gold, fill=white, line width=0.65pt]
  (4.65,1.20) rectangle (4.98,1.64);
\draw[gold!70, line width=0.45pt]
  (4.72,1.50) -- (4.91,1.50)
  (4.72,1.40) -- (4.91,1.40)
  (4.72,1.30) -- (4.87,1.30);
\node[font=\small, anchor=west, text=gold!80!black] at (5.12,1.42) {Document 4};
\node[adapter, draw=gold, fill=gold!20, text=gold!80!black] at (8.02,1.42) {+ Adapter 4};
\draw[newflow] (training.east) -- (4.43,1.42);
\node[note] at (6.75,0.38) {New documents add adapters; earlier adapters are retained.};

\node[box, draw=moss!40, minimum width=3.65cm, minimum height=0.66cm]
  (query) at (12.6,4.19) {Document + question};
\node[box, fill=mosssoft, minimum width=3.65cm, minimum height=1.05cm]
  (model) at (12.6,2.82) {Frozen language model\\[2pt]
    {\scriptsize Shared weights unchanged}};
\draw[draw=moss, fill=mosssoft, line width=0.75pt, rounded corners=0.8pt]
  (10.95,3.22) rectangle (11.24,3.44);
\draw[draw=moss, line width=0.8pt]
  (11.01,3.44) -- (11.01,3.51) arc[start angle=180,end angle=0,radius=0.085]
  -- (11.18,3.44);
\node[box, draw=moss!40, minimum width=3.65cm, minimum height=0.66cm]
  (answer) at (12.6,1.42) {Short, accurate answer};
\draw[flow] (query.south) -- (model.north);
\draw[flow] (9.25,2.82) -- (model.west);
\node[note] at (10.00,3.29) {Retrieved\\adapter};
\draw[flow] (model.south) -- (answer.north);
\end{tikzpicture}
\caption{Continual learning through locally trained document adapters. Each new document is adapted offline
on local hardware, and its adapter is added to a growing library. At serving time, the matching adapter is loaded
alongside the document and reused across queries. The base model and previously cached adapters remain unchanged.}
\label{fig:overview}
\end{figure*}
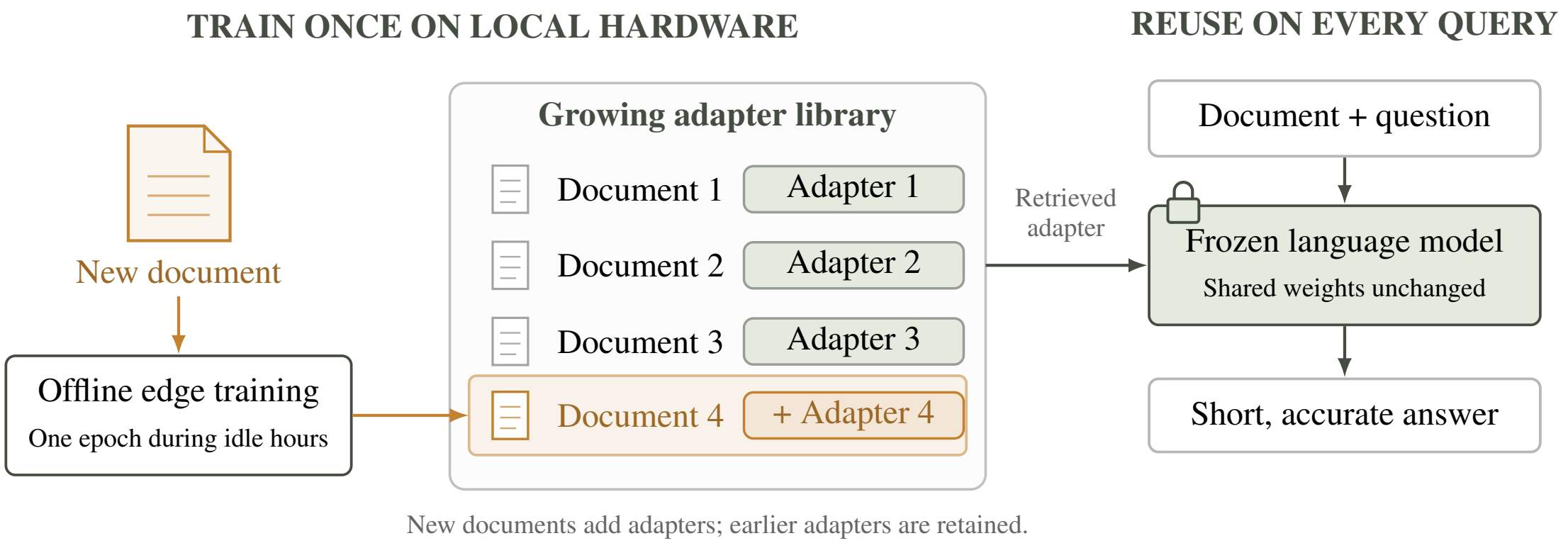

\section{Introduction}

Organizations maintain collections of private documents that grow and change over time. Specializing a language
model to such a corpus requires incorporating new information while preserving knowledge acquired from earlier
documents. Enterprise workloads offer a useful structure for this problem: users repeatedly query the same
policies, plans, specifications, and spreadsheets, and
their questions often involve recurring operations over those documents. Document-specific adaptation can therefore
be performed in advance and reused across requests, informed by both the available corpus and the questions users
ask.

Existing approaches face challenges in this setting. Fine-tuning shared model parameters can cause catastrophic
forgetting, requiring methods that balance adaptation to new data with retention of prior knowledge
\citep{kirkpatrick2017ewc,li2017lwf,wang2023olora}. Test-time training instead adapts a model to an individual input
through gradient updates \citep{sun2020ttt,akyurek2024ttt,bansal2025contextttt}. Although effective for document
question answering, performing these updates separately for each request incurs repeated computation when multiple
queries concern the same document.

We introduce \emph{microfitting}, which performs document-specific adaptation offline and caches the resulting
parameters for subsequent use (Figure~\ref{fig:overview}). For each document, we train a small low-rank adapter for one epoch on synthetic
questions in an open-book setting. The adapter is specialized to that document and loaded whenever the document is
queried. The base model remains frozen, and each document has a separate adapter, so incorporating a new document
does not overwrite parameters associated with earlier documents. Microfitting thus combines cached test-time
adaptation with parameter isolation at the document level.

Deep Context Distillation similarly trains document-level LoRA modules by matching a teacher that sees the document
\citep{caccia2025knowledgemodules}. Microfitting instead uses question-answer supervision, keeps the document in
context, and revises its training-data recipe from measured adapter outcomes.

Our method uses the Internalizer \citep{devine2026internalizer}, a document-to-adapter hypernetwork. Its initial
document-QA training and evaluation emphasized inference without the
document in context, while the original work also proposed task fine-tuning for broader downstream applications.
Here, we use the hypernetwork to initialize document-specific adapters for open-book question answering, where the document remains available at inference. With
this initialization, adapters
reach their best observed accuracy after one epoch, avoiding the additional training and checkpoint selection
required by fresh LoRA initialization. Section~\ref{sec:init} isolates this effect.

We study Enterprise Document Operations (EDO), an open-book question-answering environment over real-world
enterprise documents, including policies, planning documents, technical specifications, and tables. Because no
public benchmark closely matches this workload, we construct held-out questions with a frozen generator seeded by
examples supplied by domain experts. EDO questions require comparison, ranking, arithmetic, counting, and filtering
by multiple conditions; representative examples appear in Table~\ref{tab:examples}.

The synthetic training questions are generated by a recipe that is revised through an automatic feedback loop:
evaluate adapters against the base model, analyze corrections, regressions, and mismatches between generated and
target questions, then update the recipe for the next document batch. User questions provide examples of the
operations and forms the recipe should target. Our experiments freeze the target distribution to isolate recipe
refinement, carrying lessons from each batch of ten documents into the generation process applied to a fresh batch.
In deployment, the same update mechanism can revise the recipe as the workload distribution and its examples drift
over time. Across three EDO iterations, relative improvement over the corresponding baseline increases from 30.9\%
to 37.5\% to 38.2\% (Figure~\ref{fig:headline}).

Microfitting is designed to run entirely on customer-owned hardware, keeping documents, user questions, and the
full adaptation process within the customer's private environment.

Our contributions are:
\begin{itemize}
\item A method for document-level, parameter-isolated adaptation of a frozen 284B-parameter model using synthetic
  questions produced and revised automatically from measured outcomes.
\item A demonstration of continual improvement in EDO, where relative
  improvement over the baseline increases from 30.9\% to 37.5\% to 38.2\% as the generation recipe is revised.
\item Measurements of tradeoffs among questions per document, training epochs, document batching, and GPU
  execution, plus a workstation implementation that completes a 50-step run in 16.8 minutes.
\end{itemize}

\section{Related work}
\label{sec:related}

\paragraph{Continual learning.} Regularization methods \citep{kirkpatrick2017ewc,li2017lwf} and replay mitigate
forgetting by constraining updates to shared parameters. Adapter-based approaches constrain successive updates to
orthogonal or low-interference subspaces \citep{wang2023olora,liang2024inflora}, or use sparsity and task-specific
routing \citep{dragomir2026jumplora,azghan2026latentlora,liao2026sparta}. Parameter isolation
\citep{rusu2016progressive,mallya2018packnet,serra2018hat} assigns separate parameters to each task, preventing
subsequent updates from overwriting earlier tasks but requiring a way to select the appropriate parameters at
inference. In our setting, the document identified by the request determines which adapter to load.

Continual adaptation to document streams has been studied as a means of keeping language models current
\citep{lazaridou2021mindthegap,jang2022ckl,jin2022lifelong}. Updating on new articles
\citep{liska2022streamingqa} or changes between encyclopedia snapshots \citep{jang2022temporalwiki} can maintain
question-answering performance at lower cost than retraining, while suitable continual pre-training schedules can
match training from scratch \citep{ibrahim2024continual}. Online document adaptation has also been meta-learned
\citep{hu2023camels} and amortized into a forward pass \citep{tack2024mac}; \citet{wu2024clsurvey} provide a broader
survey. Microfitting addresses the same need to incorporate new documents through separately stored adapters,
isolating subsequent adaptation from previously cached parameters.

\paragraph{Test-time training and cached adaptation.} Test-time training adapts a model to its input through
parameter updates \citep{sun2020ttt,sun2024tttlayers,behrouz2024titans}. In open-book document question answering,
the choice of training spans can matter more than the number of update steps
\citep{bansal2025contextttt,zhu2026sttt,yuan2026easettt}. Meta-learned initializations can further reduce the number
of steps required \citep{zhang2025gradientprompting,chen2025perk,tandon2025ttte2e}. Related approaches cache the
result of adaptation: Cartridges learn a per-corpus KV cache through self-study \citep{eyuboglu2025cartridges}, and
Knowledge Modules train per-document LoRA adapters for retrieval-based serving \citep{caccia2025knowledgemodules}.
Microfitting combines open-book adaptation with persistent caching on a model an order of magnitude larger than
those used in prior work.

\begin{figure*}[t]
\centering
\begin{tikzpicture}[
  x=1cm, y=1cm,
  process/.style={hf-process},
  artifact/.style={hf-artifact},
  flow/.style={hf-flow},
  feedback/.style={hf-feedback},
  note/.style={hf-note},
  stage/.style={hf-stage}
]
\node[stage] at (7.60,4.65) {INNER LOOP $\cdot$ DOCUMENT-CONDITIONED GENERATION AND ADAPTATION};
\node[artifact, minimum width=2.10cm] (inputs) at (1.10,3.50)
  {Document $d_i$\\[-1pt]{\scriptsize $d_i\in D_t$}\\[-1pt]+ workload\\[-1pt]examples $U$};
\node[process, minimum width=2.20cm] (generate) at (3.85,3.50)
  {Synthetic-data\\[-1pt]generator\\[-1pt]{\scriptsize $G_t=G_{r_t}$}};
\node[artifact, minimum width=2.20cm] (training) at (6.65,3.50)
  {Training set\\$T_{t,i}$};
\node[process, minimum width=2.85cm] (microfit) at (9.80,3.50)
  {Adapter training $\mathcal{H}$\\[-1pt]
    {\tiny $a_{t,i}=\mathcal{H}(M_\theta,d_i,T_{t,i})$}};
\node[process, minimum width=2.45cm] (evaluate) at (13.15,3.50)
  {Evaluate on $V_{t,i}$\\[-1pt]{\scriptsize outcome $o_{t,i}$}};

\draw[flow] (inputs) -- (generate);
\draw[flow] (generate) -- (training);
\draw[flow] (training) -- (microfit);
\draw[flow] (microfit) -- (evaluate);

\node[artifact, minimum width=2.85cm] (recipe) at (3.85,1.42)
  {Data-generation policy\\[-1pt]{\scriptsize recipe $r_t$}};
\draw[flow] (recipe) -- (generate);

\node[artifact, minimum width=2.70cm] (serve) at (9.30,1.42)
  {Cached adapter $a_{t,i}$\\[-1pt]{\scriptsize serve $M_{\theta,a_{t,i}}$}};
\draw[flow] (microfit) -- (serve);

\node[stage] at (6.75,2.25) {OUTER LOOP $\cdot$ RECIPE OPTIMIZATION};
\node[process, minimum width=2.95cm] (update) at (13.15,1.42)
  {Recipe optimizer $\Phi$\\[-1pt]
    {\tiny $r_{t+1}=\Phi(r_t,U,T_t,O_t)$}};
\draw[flow] (evaluate) -- (update);
\draw[feedback] (update.south) -- (13.15,0.35) -- (4.15,0.35) -- (recipe.south);
\node[note, fill=white, inner sep=2pt] at (7.10,0.35)
  {adopt $r_{t+1}$ on measured improvement};
\end{tikzpicture}
\caption{Feedback-driven data generation and microfitting. Recipe $r_t$ configures synthetic-data generator
$G_t=G_{r_t}$, which is applied to each document $d_i$ to produce document-specific training set $T_{t,i}$.
Procedure $\mathcal{H}$ fits adapter $a_{t,i}$, which is cached for serving and evaluated on held-out questions
$V_{t,i}$. The outer-loop recipe optimizer $\Phi$ aggregates evidence across batch $D_t$ and proposes $r_{t+1}$,
which is adopted only on measured improvement.}
\label{fig:system}
\end{figure*}
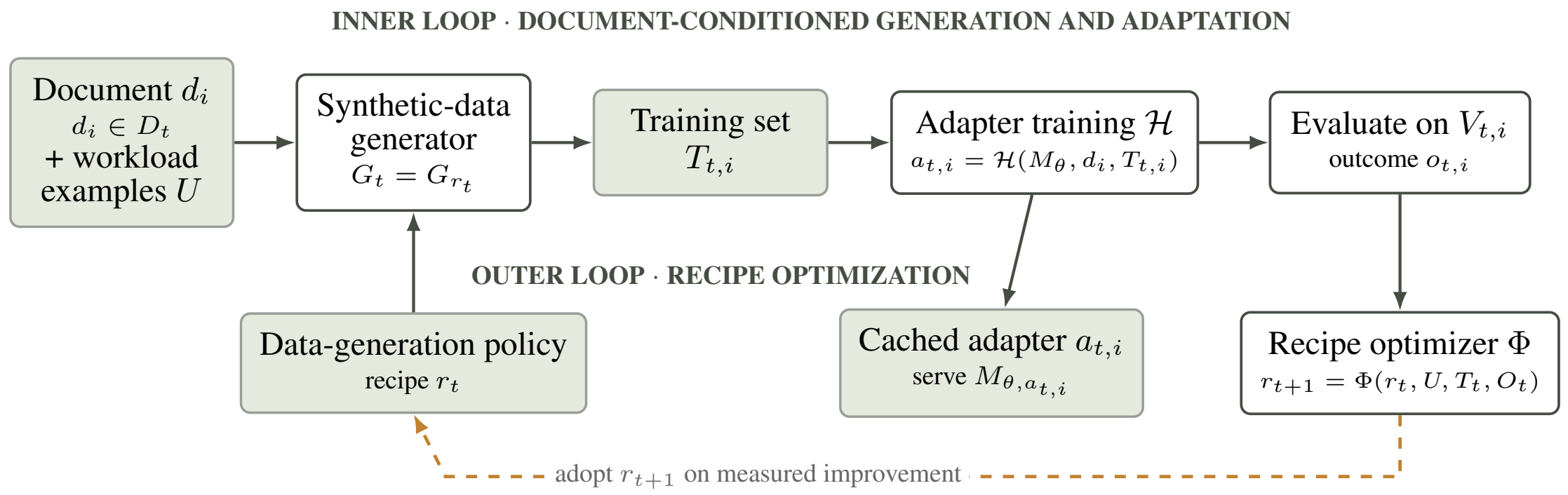

\paragraph{Context-to-parameter hypernetworks.} The Internalizer \citep{devine2026internalizer}, Generative Adapter
\citep{chen2025generativeadapter}, StreamAdapter \citep{muhtar2024streamadapter}, Doc-to-LoRA
\citep{charakorn2026doctolora}, Text-to-LoRA \citep{charakorn2025texttolora}, and DyPRAG \citep{tan2025dyprag} map
text to adapter parameters in a forward pass. We use this mapping as the starting point for further adaptation to
each document. Hypernetworks may also support transfer between base models: PorTAL
\citep{geist2026portal} recovers nearly all the gain of native adapter training by fitting a small model-specific
alignment, whereas direct adapter transfer \citep{xia2025crosslora} recovers a smaller fraction in the same
comparison.

\paragraph{Knowledge injection and synthetic question answering.} Diverse synthetic questions can improve the
acquisition of document knowledge relative to training on the document alone
\citep{allenzhu2023physics31,yang2024entigraph,zhang2024selftuning}. Retrieval remains effective when the document
is available \citep{ovadia2023finetuning}. Microfitting retains the document in the prompt and uses adaptation to
improve the model's use of its contents and structure.

\paragraph{Self-improving data pipelines.} Prior work selects data by quality or student utility
\citep{chen2023alpagasus,liu2023deita,xia2024less,li2024montessori}, generates data in response to student errors
\citep{lee2024llm2llm}, learns self-generated edits \citep{zweiger2025seal}, or employs data-generation agents in a
teacher environment \citep{khan2024dataenvgym}. Our pipeline computes reference answers with code and trains a
separate adapter for each document. Training and evaluation take only minutes, allowing the recipe to be refined
quickly based on observed performance.

\section{Microfitting}
\label{sec:method}

Microfitting converts a document collection into a library of document-specific adapters. For each batch of
documents, we generate synthetic question--answer pairs, adapt a shared working copy of the hypernetwork, and then
use it to produce one adapter per document. The adapters are cached for serving, while evaluation results are used
to improve the data-generation recipe for later batches. The base model and the initial hypernetwork checkpoint
remain fixed.

\subsection{Feedback-driven data generation and microfitting}

Figure~\ref{fig:system} separates the shared data-generation policy from its document-specific outputs. At
outer-loop iteration $t$, $D_t$ denotes the current document batch and $d_i\in D_t$ its $i$th document. Recipe
$r_t$ configures generator $G_t=G_{r_t}$; applying $G_t$ to $d_i$ produces a document-specific training set. The
remainder of the inner loop fits and evaluates an adapter and caches it for serving. The outer loop aggregates
evidence across the batch and revises the recipe that configures the next generator.

The nested data-generation and adaptation processes can be written compactly as
\begin{equation}
\begin{aligned}
G_t &\equiv G_{r_t}, \\
T_{t,i} &\sim G_t(\,\cdot \mid d_i,U), \\
a_{t,i} &= \mathcal{H}(M_\theta,d_i,T_{t,i}), \\
o_{t,i} &= \operatorname{Eval}(M_{\theta,a_{t,i}},V_{t,i}), \\
r_{t+1} &= \Phi(r_t,U,T_t,O_t).
\end{aligned}
\label{eq:system}
\end{equation}
Here, $r_t$ is the data-generation policy, implemented as a recipe of prompts and constraints, and $U$ contains
examples of the target workload. In production, $U$ would ordinarily consist of historical questions from real
users. The training generator $G_t$ conditions on $d_i$ and $U$ to produce synthetic training set $T_{t,i}$. The
held-out set $V_{t,i}$ is obtained separately and is never produced by $G_t$ or used for adapter training. In
production, $V_{t,i}$ would be a reserved, scored set of real-user questions about $d_i$.
Let $\theta$ denote the frozen base parameters and let
$a_{t,i}=\{(A_{\ell,t,i},B_{\ell,t,i})\}_{\ell\in\mathcal{L}}$ collect the LoRA factors for the
adapted layers. Following standard LoRA notation \citep{hu2022lora}, the effective weight at adapted layer $\ell$ is
$W_{\ell,t,i}=W_{\ell,0}+(\alpha/r)B_{\ell,t,i}A_{\ell,t,i}$; we denote the resulting model by
$M_{\theta,a_{t,i}}$. The microfitting procedure $\mathcal{H}$ produces adapter $a_{t,i}$ for $M_\theta$, and
$o_{t,i}$ records its measured outcomes on $V_{t,i}$. Writing $T_t=\{T_{t,i}:d_i\in D_t\}$ and
$O_t=\{o_{t,i}:d_i\in D_t\}$, the outer-loop
recipe optimizer $\Phi$ analyzes the batch-level evidence to produce the next recipe. $\Phi$ does not generate
examples or update model weights directly; it updates $r_t$, which configures the next generator $G_{t+1}$.

The two loops need not operate at the same document granularity. In the reported experiment, $\Phi$ updates the
recipe after each ten-document batch, but it could instead aggregate evidence from multiple completed runs before
proposing $r_{t+1}$. This would allow the outer loop to learn from a larger document sample without increasing the
inner-loop training batch size.

In the evaluated configuration, $G_t$ produces forty questions and reference answers for each document. Procedure
$\mathcal{H}$ initializes an adapter with the hypernetwork and adapts it for one epoch on $T_{t,i}$. The resulting
$a_{t,i}$ is exported as a standard LoRA checkpoint and loaded for later requests about $d_i$. Both adaptation and
inference are open-book: the document remains in the prompt, and $a_{t,i}$ improves how $M$ uses it.

\subsection{Base model, adapter and hypernetwork}
\label{sec:hypernet}

The base model is DeepSeek V4 Flash \citep{deepseekv4flash}, a 284B-parameter mixture-of-experts model whose
parameters remain frozen. Each adapter applies rank-16 LoRA \citep{hu2022lora} to the three
shared-expert projections in every layer (Table~\ref{tab:hparams}).

Procedure $\mathcal{H}$ initializes the adapters using the Internalizer \citep{devine2026internalizer}, a
hypernetwork that maps document $d$ to LoRA parameters for base model $M$. Before the microfitting experiments, we
further trained the hypernetwork on extremely simple retrieval Q\&A over synthetic enterprise-style documents to
move it into a general Q\&A regime. Microfitting then adapts that checkpoint rapidly to the more complex, realistic
document-Q\&A tasks studied here (Appendix~\ref{app:hypernet}). During $\mathcal{H}(M,d_i,T_{t,i})$, the answer-token loss is
backpropagated through the generated adapter, and the optimizer updates the hypernetwork's parameters. After one
epoch, the resulting $a_{t,i}$ is exported as a standard LoRA checkpoint. The working copy of the
hypernetwork is then discarded; each new batch starts from the same initial checkpoint. The hypernetwork is used
only during training, and the exported adapters can be loaded by a serving stack that supports standard LoRA
checkpoints.

The Internalizer supports both closed-book document Q\&A, where the adapter replaces the document in the prompt,
and open-book Q\&A, where the document remains available in context. Here, we use its hypernetwork to initialize
document-specific adapters for the open-book setting. Training across ten documents enables transfer within the
batch, while conditioning the trained hypernetwork on each document produces a distinct adapter for
document-specific adaptation.

The hypernetwork provides the initialization used by the fixed adaptation schedule; Section~\ref{sec:init} compares
its convergence with fresh LoRA.

Separately, a small model-specific refit can transfer a hypernetwork to a new base model
\citep{geist2026portal}; our internal demonstrations of such transfer support regenerating the adapter collection
after a base-model upgrade using the same offline adaptation process.

\subsection{Training configuration}
\label{sec:recipe}

\begin{table}[t]
\centering
\footnotesize
\begin{tabular}{@{}ll@{}}
\toprule
base model & DeepSeek V4 Flash, frozen \\
adapter & LoRA rank 16, shared-expert MLP \\
trainable parameters & 12.7M per adapter \\
documents per batch & 10 \\
questions per document & 40 \\
rows per step & 8, one question each \\
steps per epoch & 50 \\
epochs & 1 \\
loss & cross-entropy on answer tokens \\
optimizer, matrices & Muon, $3\times10^{-4}$ \\
optimizer, other & AdamW, $2\times10^{-5}$ \\
schedule & 5 warm-up steps, then constant \\
\bottomrule
\end{tabular}
\caption{Default training configuration, held fixed across question-generation recipe revisions.}
\label{tab:hparams}
\end{table}

The training configuration is designed for scheduled execution on enterprise-owned hardware
(Table~\ref{tab:hparams}). Each $D_t$ contains ten documents, with forty questions generated per document. With
eight question--answer examples per step, one epoch comprises fifty steps. Every prompt includes $d$,
and
cross-entropy loss is applied only to answer tokens. We use Muon \citep{jordan2024muon,liu2025muonscalable} for
matrix parameters and AdamW for the remaining parameters, without per-document or per-batch tuning.

Although $\mathcal{H}$ updates a shared hypernetwork while processing $D_t$, each $d_i$ retains its own $a_{t,i}$.
After training, each document is passed through the hypernetwork to generate its document-specific parameters; the
shared checkpoint is not itself a single LoRA applied to every document in the batch.
This design combines transfer across the ten-document batch with document-level specialization, so each cached
adapter benefits from shared training while remaining tailored to its own document.

We use forty questions per document as a tested operating point based on the sparse data-efficiency sweep in
Section~\ref{sec:efficiency}. The sweep did not evaluate values between forty and approximately eighty-five and
therefore does not establish a global optimum; the best setting may depend on the corpus, question recipe, and
hypernetwork initialization. Separate batching experiments compare individual and joint training; the batch
composition used elsewhere was an informed design choice rather than the result of an exhaustive search.

The one-epoch schedule reflects observed generalization performance. Across the two held-out sets used to choose
the schedule, a second epoch did not improve accuracy: it slightly reduced accuracy on one and left it unchanged on
the other. Continued optimization lowered training loss but did not improve performance on new questions,
indicating a closer fit to the training questions.

\subsection{Parameter isolation and continual learning}
\label{sec:isolation}

Parameter isolation preserves previously acquired knowledge as documents are added. The base model $M$ remains
unchanged, and each cached $a_{t,i}$ is stored separately, so training on new documents does not modify parameters
used to serve earlier documents. Individual adapters can also be regenerated or deleted independently, enabling
the refresh process described in Section~\ref{sec:loop}. Candidate $r_{t+1}$ is accepted only after a paired
comparison on a fresh validation subset: the incumbent and candidate recipes each generate training data and fit
adapters for the same documents, and the resulting adapters are scored on the same new held-out questions. These
documents and questions did not inform the candidate recipe. Otherwise, $r_t$ and the existing cached adapters are
retained.
Although documents in the same training batch share
hypernetwork updates, the batching ablations show no reduction in per-document accuracy relative to individual
training.

This construction supports continual adaptation to a changing corpus: new documents receive new adapters, revised
documents receive replacement adapters, and recipe improvements can be applied to existing documents. Our
EDO demonstration evaluates successive recipe versions on fresh document
samples while holding the target question generator fixed. We have not yet measured adaptation under a changing
user-question distribution (Section~\ref{sec:compounding}).

\section{Data-generation policy refinement}
\label{sec:loop}

This section describes how the data-generation recipe improves from measured adapter outcomes. The outer loop
analyzes corrections, regressions, and mismatches with the target workload, then adopts revised recipes only when
held-out performance improves.

\subsection{Iterative recipe optimization}

Recipe $r_t$ consists of a prompt and constraints for producing questions with reference answers. It configures
$G_t=G_{r_t}$, which is applied to each $d_i\in D_t$ to produce $T_{t,i}$. Procedure $\mathcal{H}$ fits $a_{t,i}$,
and evaluation on $V_{t,i}$ produces outcome $o_{t,i}$: a correction, regression, or persistent error relative to
$M$. The outer-loop recipe optimizer $\Phi(r_t,U,T_t,O_t)$ proposes a single candidate
$r_{t+1}$ with an explicit hypothesis. Acceptance uses a paired test on a fresh document subset: $r_t$ and
$r_{t+1}$ each generate training sets and fit separate adapters for the same documents, and both sets of adapters
are evaluated on the same new held-out questions. These questions are separate from $V_t$, whose outcomes already
helped shape $r_{t+1}$. The candidate is adopted only if it improves over the incumbent under the same fixed
training configuration and evaluation protocol.

Matching the empirical distribution of $U$ exactly is not itself the objective. $U$ anchors the target workload,
but $\Phi$ may deliberately rebalance $T_t$ toward question types that the base model misses and document-specific
adapters can correct, provided that the change improves performance on held-out workload questions. The
correctable-error study explores this distinction between imitating the workload and allocating training to model
deficiencies.

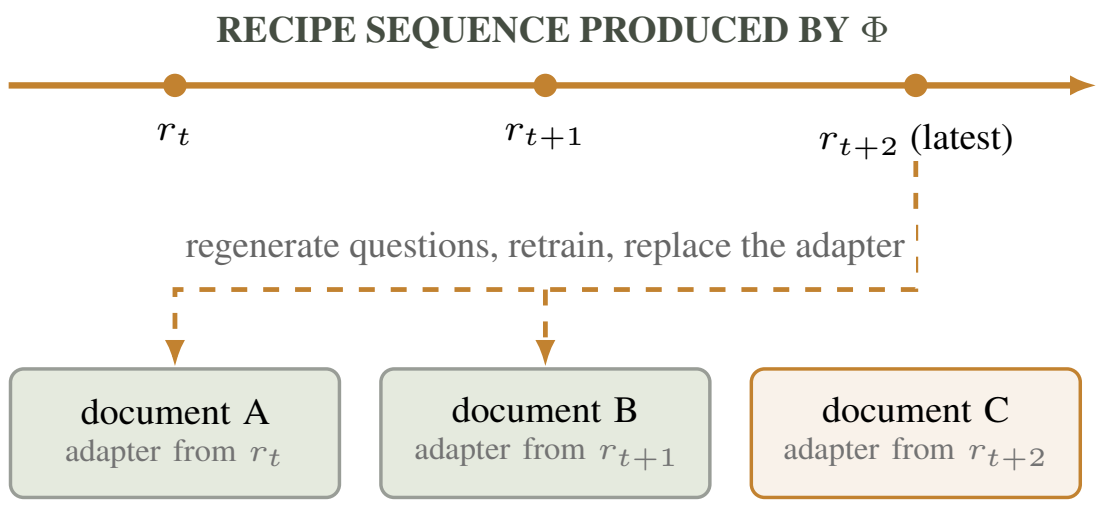
\begin{figure}[t]
\centering
\begin{tikzpicture}[x=1cm, y=1cm,
  n/.style={hf-artifact, font=\scriptsize, inner sep=2.5pt,
            text width=2.0cm, minimum height=8.5mm},
  egold/.style={hf-feedback},
  el/.style={hf-note, inner sep=1pt}
]
\draw[gold, line width=1.2pt, -{Latex[length=1.8mm]}] (0.1,0) -- (7.3,0);
\node[hf-stage] at (3.7,0.35) {RECIPE SEQUENCE PRODUCED BY $\Phi$};
\foreach \x/\l in {1.2/{$r_t$}, 3.65/{$r_{t+1}$}, 6.1/{$r_{t+2}$ (latest)}} {
  \fill[gold] (\x,0) circle (2.2pt);
  \node[font=\scriptsize, anchor=north] at (\x,-0.12) {\l};
}
\node[n] (a) at (1.2,-2.3) {document A\\[-1pt]{\tiny\color{black!55}adapter from $r_t$}};
\node[n] (b) at (3.65,-2.3) {document B\\[-1pt]{\tiny\color{black!55}adapter from $r_{t+1}$}};
\node[n, draw=gold, fill=gold!10] (c) at (6.1,-2.3)
  {document C\\[-1pt]{\tiny\color{black!55}adapter from $r_{t+2}$}};
\draw[egold] (6.1,-0.5) -- (6.1,-1.35) -- (1.2,-1.35) -- (a.north);
\draw[egold] (3.65,-1.35) -- (b.north);
\node[el, fill=white, inner sep=2pt] at (3.65,-1.08) {regenerate questions, retrain, replace the adapter};
\end{tikzpicture}
\caption{Adapter refresh after $\Phi$ adopts a recipe update. During idle periods, the latest generator
$G_{t+1}=G_{r_{t+1}}$ regenerates document-specific training sets for existing documents, $\mathcal{H}$ retrains
their adapters, and the system replaces the cached checkpoints.}
\label{fig:fleet}
\end{figure}

Figure~\ref{fig:system} separates two forms of learning. The inner loop
$G_t\rightarrow\mathcal{H}\rightarrow\operatorname{Eval}$ generates training sets and fits and measures document
adapters. The outer-loop optimizer $\Phi$ reviews $r_t$, workload examples $U$, training sets $T_t$, outcomes
$O_t$, and measured mismatches to propose the policy for the next generator. After $r_{t+1}$ is adopted, idle
compute capacity can use $G_{t+1}$ to regenerate document-specific training sets and replace older adapters,
progressively updating the corpus to the latest recipe (Figure~\ref{fig:fleet}).

\begin{table}[t]
\centering
\footnotesize
\setlength{\tabcolsep}{2pt}
\begin{tabular}{@{}clrrr@{}}
\toprule
iteration & policy & \shortstack{baseline $\to$\\Q\&A adapter} &
\shortstack{gain\\(points)} & \shortstack{relative\\improvement} \\
\midrule
1 & $r_0$ & $.485 \to .635$ & $+15.0$ & $+30.9\%$ \\
2 & $r_1$ & $.480 \to .660$ & $+18.0$ & $+37.5\%$ \\
\rowcolor{mosssoft} \textbf{3} & $\mathbf{r}_{2}$ & $\mathbf{.510 \to .705}$ & $\mathbf{+19.5}$ & $\mathbf{+38.2\%}$ \\
\bottomrule
\end{tabular}
\caption{Automatic refinement in EDO. Each $r_t$
is evaluated on a fresh held-out sample of ten documents and 200 questions from the same frozen target distribution.
Relative improvement is computed against the baseline from the same iteration.}
\label{tab:prospective}
\end{table}

\subsection{Automatic refinement in EDO}

We demonstrate refinement in EDO, where the base model struggles with many of the target questions. Because we do
not have a large production query log for this study,
we simulate one with a frozen target-question generator seeded by examples supplied by domain experts. This
generator produces workload examples and document-specific held-out questions from a fixed target distribution. It
is separate from $G_t$, whose recipe is refined to generate adapter-training data. Although $G_t$ conditions on
workload examples $U$, it receives neither $V_{t,i}$ nor the prompt or procedure used by the frozen target-question
generator. Training and evaluation questions are therefore produced by separate processes. The resulting simulated
workload spans comparison, counting, arithmetic, conditional filtering, ranking, and lookup questions. The method is not
specific to this environment: a deployment can continue updating the policy as documents and workload examples
change.

We construct the initial recipe $r_0$ from target questions and base-model outcomes on a three-document sample. We
evaluate $r_0$ on a fresh held-out batch $D_0$ of ten documents. The next update receives the resulting $T_0$ and
$O_0$ together with measurements of how the generated questions differ from the target workload $U$. These
differences include set size, wording, whether the relevant group is named, whether counting conventions are stated,
and the requested answer form. Applying $\Phi(r_0,U,T_0,O_0)$ produces $r_1$, which is evaluated on another fresh
batch $D_1$. Applying the same update to $(r_1,U,T_1,O_1)$ produces $r_2$, which is evaluated on fresh batch $D_2$.

A separate correctable-error analysis with changing question distributions across recipes appears in
Section~\ref{sec:correctable-errors}, Table~\ref{tab:correctable-errors}, and Figure~\ref{fig:recipes}.

\subsection{System-discovered task slices}
\label{sec:kinds}

The outer-loop optimizer $\Phi$ does not begin with a fixed taxonomy of tasks. It defines categories and performance
slices from workload examples $U$, generated sets $T_t$, and measured outcomes $O_t$, then revises those slices as
new evidence arrives. In the correctable-error targeting experiment, for example, it found that $M$ already
performed well on direct retrieval tasks, including single-fact and rule lookups and list-membership questions, and
identified three more challenging slices. \emph{Conditional-comparison} questions require a comparison or condition over a small named
set, such as identifying the team associated with the most rows, the longest deadline, or the person appearing in
two lists.
\emph{Named-fact arithmetic} combines two or three numbers from the document in a single operation, while
\emph{counts} ask how many items satisfy a condition.

\subsection{Initializing the data-generation process}
\label{sec:quality}

We initialize $G_0$ with general priors for constructing reliable training data rather than an environment-specific
taxonomy or finished recipe. These priors include computing reference answers from parsed document representations
and code when possible, independently solving questions to check answer agreement, and using agreement and realism
gates to trigger revision or exclusion. They provide stable quality controls while $\Phi$ learns the task slices,
generation policy, and environment-specific constraints from $U$, $T_t$, and $O_t$.

\begin{figure}[t]
\centering
\begin{tikzpicture}
\begin{axis}[
  width=0.97\columnwidth, height=4.6cm,
  xlabel={Recipe refinement stage}, ylabel={Improvement over baseline (\%)},
  xmin=-0.2, xmax=2.2, ymin=0, ymax=65,
  xtick={0,1,2}, xticklabels={$r_0$,$r_1$,$r_2$},
  ytick={0,10,20,30,40,50,60},
  axis lines=left, tick align=outside,
  ymajorgrids, grid style={black!12}, axis line style={black!55},
  tick label style={font=\scriptsize}, label style={font=\footnotesize}
]
\addplot[color=plotgreen, line width=1.1pt, mark=*, mark size=2.4pt,
  mark options={solid,fill=plotgreen}]
  coordinates {(0,9.9) (1,42.7) (2,55.6)};
\node[font=\scriptsize, anchor=south, yshift=4pt] at (axis cs:0,9.9) {9.9\%};
\node[font=\scriptsize, anchor=south, yshift=4pt] at (axis cs:1,42.7) {42.7\%};
\node[font=\scriptsize, anchor=south east, xshift=3pt, yshift=4pt] at (axis cs:2,55.6) {55.6\%};
\end{axis}
\end{tikzpicture}
\caption{Relative improvement under targeted recipe refinement. Each point compares microfit accuracy with
the corresponding base model on the held-out distribution associated with $r_t$. Because the distribution changes
between stages, the line represents increasingly targeted adaptation opportunities, not a controlled accuracy
trend on a fixed workload.}
\label{fig:recipes}
\vspace{0.75em}
\centering
\footnotesize
\setlength{\tabcolsep}{3pt}
\begingroup
\renewcommand{\arraystretch}{1.18}
\begin{tabular}{@{}>{\raggedright\arraybackslash}p{2.8cm}rrrr@{}}
\toprule
recipe focus & base & microfit & \shortstack{gain\\(points)} & \shortstack{\% over\\base} \\
\midrule
$r_0$: broad initial mix & .669 & .735 & 6.6 & 9.9\% \\
$r_1$: more base-model failures & .361 & .515 & 15.4 & 42.7\% \\
\rowcolor{mosssoft} $\mathbf{r}_{2}$: \textbf{more adapter-correctable failures} & \textbf{.399} & \textbf{.621} & \textbf{22.2} & \textbf{55.6\%} \\
\bottomrule
\end{tabular}
\endgroup
\captionof{table}{Exact values underlying Figure~\ref{fig:recipes}. Each recipe is evaluated on its associated held-out
question distribution over the same ten documents. Gain is measured in percentage points, and relative improvement
is computed against the corresponding base-model accuracy.}
\label{tab:correctable-errors}
\end{figure}

\section{Experimental setup}
\label{sec:setup}

\paragraph{Enterprise Document Operations.} EDO is built from real-world enterprise documents and questions. The documents combine internal enterprise
materials with publicly available enterprise documents, and the target-question generator is based on sample
questions supplied by domain experts. Its three completed held-out iterations each contain ten previously unused,
table-rich documents and 200 questions
from the same frozen target generator. The initial $r_0$ is built from a separate three-document sample. Update
$\Phi(r_0,U,T_0,O_0)$ uses evidence from the first held-out iteration to produce $r_1$, which is evaluated on the
second; evidence from that iteration similarly produces $r_2$, evaluated on the third.

\paragraph{Correctable-error study.} A separate study tests whether the outer loop can identify question types that
the base model answers incorrectly but document-specific adapters can learn, then revise the generation recipe to
emphasize those opportunities. Unlike the main refinement experiment, the target question distribution changes
across recipe versions, so this study measures adaptive targeting rather than improvement on a fixed workload. It
uses ten documents ranging from 2.2K to 4K tokens, covering policies, planning documents, technical specifications,
and tables. Each document has forty training questions and approximately twenty evaluation questions. The three
recipes use the same documents but different training and held-out questions, increasingly concentrated
on correctable base-model deficiencies.

\paragraph{Protocol.} All reported accuracies are measured on questions excluded from adapter training. To limit
computational cost, we disable the base model's thinking mode during training and evaluation.
All systems receive the same prompt, byte for byte, with no refusal instructions; questions are submitted
individually using greedy decoding and a 128-token output limit. Claude Sonnet~5 judges each response against its
reference answer. The lead author also performed periodic human spot checks, reviewing generated questions and
reference answers for data quality and inspecting model outputs alongside the LLM-as-judge labels for grading
consistency. The EDO demonstration uses one base-model run per held-out sample; the correctable-error
study averages three runs because the distributed inference backend is not bitwise deterministic, even with greedy
decoding. An adapter
\emph{corrects} an error when it answers a question that the base model missed and \emph{regresses} when it fails a
question that the base model answered correctly.

\section{Results}
\label{sec:results}

We first evaluate whether microfitting improves accuracy on fresh documents. We then examine which errors are
correctable, how question coverage and document batching affect performance, the cost and latency of adaptation,
and the value of hypernetwork initialization.

\subsection{Demonstration: automatic improvement on fresh documents}

Table~\ref{tab:prospective} shows continual improvement across the three completed iterations. Adapter gains
increase on each fresh document batch as the recipe changes. In a supporting ablation, the revised recipe also
outperformed the preceding recipe when both were applied to the same new held-out batch. This comparison separates
improvements to the recipe from differences among evaluation batches.

\subsection{Targeted adaptation to correctable model errors}
\label{sec:correctable-errors}

The separate correctable-error study asks whether the outer loop can identify increasingly difficult but still
correctable question types---cases the base model misses but document-specific adapters can learn---and concentrate
generation on them. Table~\ref{tab:correctable-errors} and Figure~\ref{fig:recipes} show the largest gains when
generation targets errors that the adapters can correct. The same procedure could be used to build customized training
corpora, including future whole-model fine-tuning corpora, that allocate training to weaknesses observed in a
particular data environment.

We describe the stages retrospectively as moving from a broad initial mix toward base-model failures and then
adapter-correctable failures. These are summaries of the question distributions produced by $r_0$, $r_1$, and
$r_2$, not explicit recipe labels or single optimization targets; each revision changed several generation
constraints in response to measured outcomes.

Because each recipe has its own held-out distribution, these gains do not measure performance on a fixed workload.
They measure whether the system can find and target increasingly difficult, correctable slices.

\subsection{Question coverage and document batching}
\label{sec:efficiency}

\begin{figure}[t]
\centering
\begin{tikzpicture}
\begin{axis}[
  width=\columnwidth, height=5.8cm,
  xlabel={Training questions per document}, ylabel={Accuracy (\%)},
  xmin=4, xmax=93, ymin=.55, ymax=.80,
  xtick={12,24,40,85}, ytick={.55,.60,.65,.70,.75,.80},
  yticklabels={55,60,65,70,75,80},
  axis lines=left, tick align=outside,
  ymajorgrids, grid style={black!12}, axis line style={black!55},
  tick label style={font=\footnotesize}, label style={font=\footnotesize},
  legend style={at={(0.5,-0.30)}, anchor=north, draw=none, fill=none,
    font=\footnotesize, cells={anchor=west}, row sep=2pt},
  legend cell align=left
]
\addplot[color=plotgreen, line width=1.1pt, mark=*, mark size=2.4pt,
  mark options={solid,fill=plotgreen}, forget plot]
  coordinates {(12,.632) (24,.696) (40,.735) (85,.725)};
\addplot[color=basegrey, thick, dashed, domain=4:93, forget plot] {.598};
\addlegendimage{color=plotgreen, line width=1.1pt, mark=*, mark size=2.4pt}
\addlegendentry{Q\&A adapter; average time per document}
\addlegendimage{color=basegrey, thick, dashed}
\addlegendentry{Base model}
\node[font=\scriptsize, anchor=north, yshift=-5pt] at (axis cs:12,.632) {0.44 min};
\node[font=\scriptsize, anchor=south, yshift=5pt] at (axis cs:24,.696) {0.90 min};
\node[font=\scriptsize, anchor=south, yshift=5pt] at (axis cs:40,.735) {1.48 min};
\node[font=\scriptsize, anchor=south, yshift=5pt] at (axis cs:85,.725) {3.17 min};
\end{axis}
\end{tikzpicture}
\caption{Training question coverage. Accuracy is measured after one epoch on 204 evaluation questions
generated under $r_0$, with ten documents per batch. Time annotations are average training time per document on
eight H100s, computed by dividing batch time by ten.}
\label{fig:efficiency}
\end{figure}
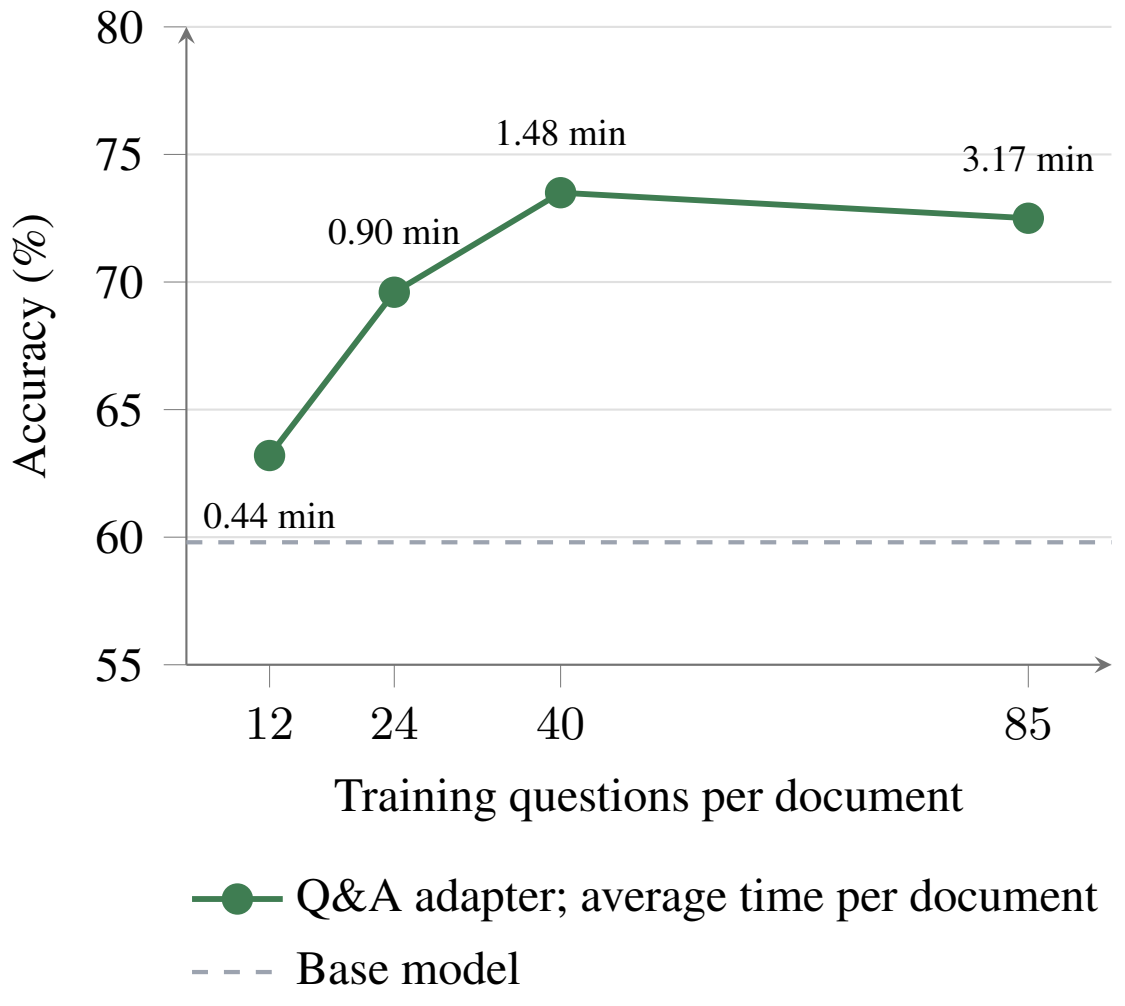

\begin{table}[t]
\centering
\small
\begin{tabular}{@{}rrr@{}}
\toprule
questions per document & accuracy & \shortstack{average time\\per document} \\
\midrule
12 & .632 & 0.44 min \\
24 & .696 & 0.90 min \\
\rowcolor{mosssoft} \textbf{40} & \textbf{.735} & \textbf{1.48 min} \\
85 & .725 & 3.17 min \\
\bottomrule
\end{tabular}
\caption{Question-count ablation with ten documents and one training epoch. All runs are evaluated on the
same 204 questions; base-model accuracy is .598. Average per-document times are batch times divided by ten and
were measured on eight H100s.}
\label{tab:question-count}
\end{table}

Figure~\ref{fig:efficiency} and Table~\ref{tab:question-count} show gains from twelve to forty questions per
document, while the approximately eighty-five-question condition is slightly worse and more than twice as slow.
Because the sweep does not test the range from fifty to seventy questions, it cannot determine whether performance
peaks at forty, somewhere between forty and eighty-five, or earlier. We therefore use forty as a reasonable tested
operating point, not as an optimum. Table~\ref{tab:document-batching} provides only an indicative comparison: adding
a large corpus of general documents to the ten-document batch did not improve accuracy. We use ten as a practical
operating point and leave optimization of batch size and composition to future work.

\begin{table}[h]
\centering
\footnotesize
\setlength{\tabcolsep}{3pt}
\begin{tabular}{@{}lrrrrr@{}}
\toprule
training condition & 1 & 2 & 3 & 10 & \shortstack{10 +\\general} \\
\midrule
evaluation accuracy & .697 & .738 & .775 & \textbf{.835} & \textbf{.835} \\
\bottomrule
\end{tabular}
\caption{Individual versus batched training in an earlier experiment. Accuracy is averaged over the same two
evaluation documents after one epoch. For each document, the first four conditions train it alone, with the other
evaluation document, in a separate three-document group, or in the full ten-document batch. The final condition
adds a large corpus of general documents.}
\label{tab:document-batching}
\end{table}

\subsection{Cost, speed and latency}
\label{sec:cost}

The private-deployment case for microfitting need not assume a datacenter-class cluster: the workload can be
scaled down to an edge/workstation-scale system built from more accessible hardware. We demonstrate this by
optimizing the same workload for two RTX PRO 6000 workstation GPUs, with model weights offloaded to system RAM.
Reusing document encodings, overlapping weight transfer with compute, and sharing document prefixes across
questions together reduced a 50-step run from nearly two hours to 16.8 minutes, a cumulative
$6.7\times$ speedup (Table~\ref{tab:workstation-optimization}). The optimized workstation and H100 runs follow the
same loss trajectory, indicating that these systems changes preserve the training behavior. At this throughput,
scheduled overnight adaptation can process several hundred documents on a single two-GPU workstation.

With thinking disabled, microfitted responses have a median length of three output tokens, compared with 378 tokens
for the same base model with thinking enabled (Appendix~\ref{app:latency}).

\begin{table}[t]
\centering
\small
\begin{tabular}{@{}lrrr@{}}
\toprule
\shortstack[l]{cumulative\\configuration} & \shortstack{warm\\s/step} & \shortstack{50-step\\time} & speedup \\
\midrule
baseline (RAM offload) & 134 & $\sim$112 min & $1.0\times$ \\
cache + 2 q./row & $\sim$57 & $\sim$52 min & $2.2\times$ \\
windowing + shared prefix & $\sim$36 & $\sim$32 min & $3.5\times$ \\
\rowcolor{mosssoft} \textbf{4 q./row + sharing} & \textbf{18.3} & \textbf{16.8 min} & $\mathbf{\sim 6.7\times}$ \\
\bottomrule
\end{tabular}
\caption{Cumulative optimization of a 50-step training run on two RTX PRO 6000 workstation GPUs.
Speedup is calculated from total 50-step training time. Intermediate rows include the preceding optimizations; the
final configuration measures 22.8 seconds per step while filling the encoding cache and 18.3 seconds with a warm
cache. Warm seconds per step reports steady-state throughput and is not used to calculate full-run speedup.}
\label{tab:workstation-optimization}
\end{table}

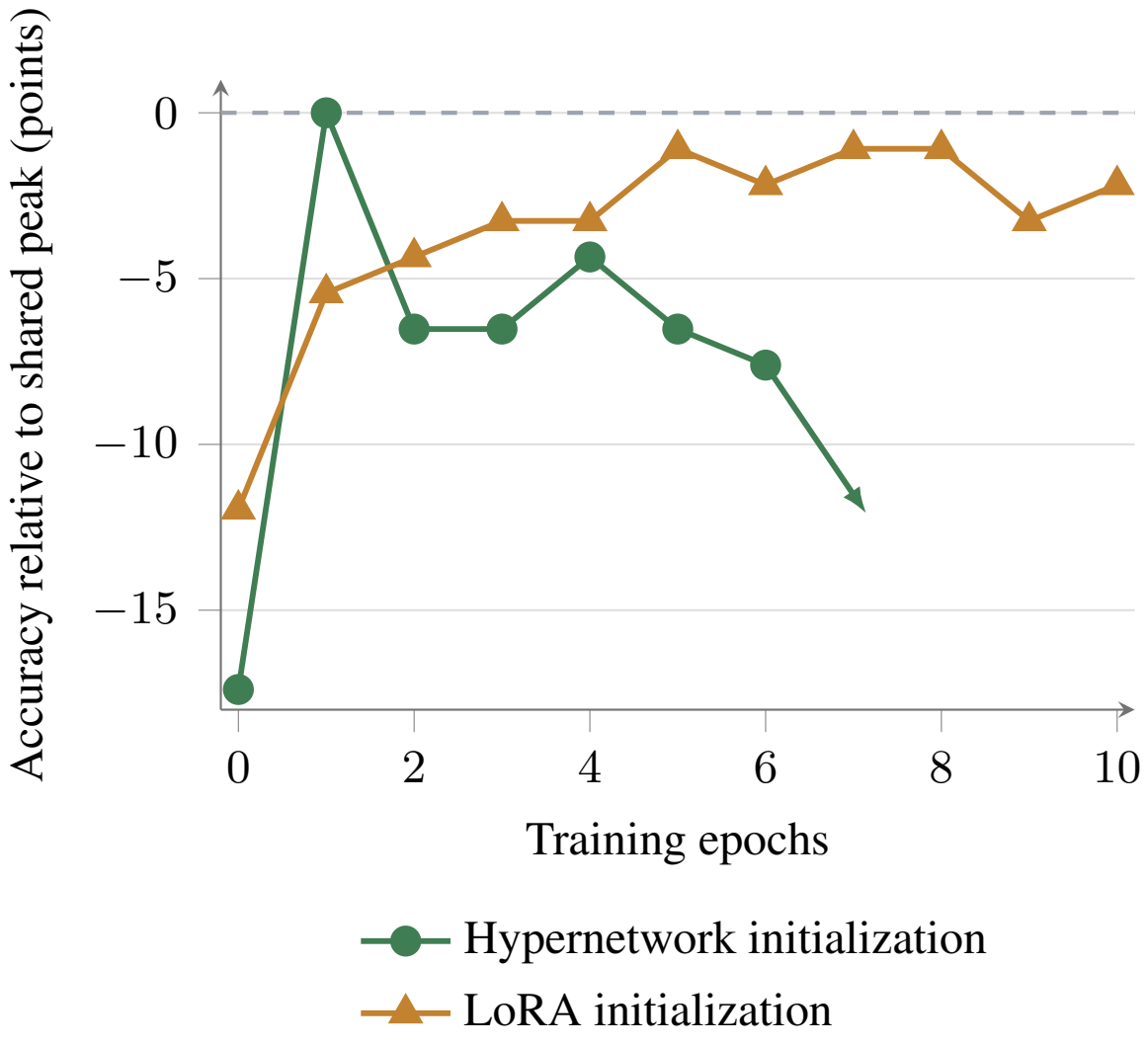
\begin{figure}[t]
\centering
\begin{tikzpicture}
\begin{axis}[
  width=\columnwidth, height=5.8cm,
  xlabel={Training epochs}, ylabel={Accuracy relative to shared peak (points)},
  xmin=-0.2, xmax=10.2, ymin=-18, ymax=1,
  xtick={0,2,4,6,8,10}, ytick={-15,-10,-5,0},
  axis lines=left, tick align=outside,
  ymajorgrids, grid style={black!12}, axis line style={black!55},
  tick label style={font=\footnotesize}, label style={font=\footnotesize},
  legend style={at={(0.5,-0.30)}, anchor=north, draw=none, fill=none,
    font=\footnotesize, cells={anchor=west}, row sep=2pt},
  legend cell align=left
]
\addplot[color=plotgreen, line width=1.1pt, mark=*, mark size=2.4pt,
  mark options={solid,fill=plotgreen}]
  coordinates {(0,-17.3913) (1,0) (2,-6.5217) (3,-6.5217) (4,-4.3478)
    (5,-6.5217) (6,-7.6087)};
\addlegendentry{Hypernetwork initialization}
\draw[-{Latex[length=1.8mm]}, color=plotgreen, line width=1.1pt]
  (axis cs:6,-7.6087) -- (axis cs:7.15,-12.1);
\addplot[color=gold, line width=1.1pt, mark=triangle*, mark size=2.8pt,
  mark options={solid,fill=gold}]
  coordinates {(0,-11.9565) (1,-5.4348) (2,-4.3478) (3,-3.2609) (4,-3.2609)
    (5,-1.0870) (6,-2.1739) (7,-1.0870) (8,-1.0870) (9,-3.2609) (10,-2.1739)};
\addlegendentry{LoRA initialization}
\addplot[color=basegrey, thick, dashed, domain=-0.2:10.2, forget plot] {0};
\end{axis}
\end{tikzpicture}
\caption{Initialization comparison on a separate development example. Accuracy is measured relative to the shared
best checkpoint across both runs. The hypernetwork peaks after one epoch; fresh LoRA stays below it through ten
epochs.}
\label{fig:epochs}
\end{figure}

\subsection{Hypernetwork initialization and convergence}
\label{sec:init}

Because Gemma 4 26B \citep{gemma2026gemma4} is smaller, we use it for higher-volume controlled ablations on a
separate training set. In Figure~\ref{fig:epochs}, hypernetwork initialization peaks after one epoch; fresh LoRA
remains below it through epoch ten and still requires checkpoint selection. We observe the same convergence pattern
on DeepSeek V4 Flash: although hypernetwork initialization and fresh LoRA reach similar peak accuracy, hypernetwork
initialization reaches its best observed performance after one epoch, whereas fresh LoRA generally requires five or
six epochs and checkpoint selection. This agrees with HyperTuning \citep{phang2022hypertuning}, which
found faster convergence and better final performance from hypernetwork-generated parameters than from random or
shared adapter initializations.

\section{Error analysis and interpretation}
\label{sec:analysis}

Aggregate accuracy does not show which errors microfitting corrects or where adaptation stops helping. We
examine representative corrections across increasing numbers of relevant items, then interpret what the adapters
encode.

\subsection{Worked examples}

\begin{table*}[t]
\centering
\small
\begin{tabular}{>{\raggedright\arraybackslash}p{9.3cm}
                >{\raggedright\arraybackslash}p{2.6cm}
                >{\raggedright\arraybackslash}p{2.8cm}}
\toprule
question & base model & gold \\
\midrule
In the Home Office immigration and nationality fees: 9 April 2025 schedule, a sponsor applies in the UK for a
worker holding a four-year certificate of sponsorship for a job on the Immigration Salary List that is not a
Health and Care Visa job. Using section 7 (Work -- applications made in the UK), how much more would that
application cost from 9 April 2025 if the job were not on the Immigration Salary List? &
\pounds 115 & \pounds 591 \\
\addlinespace
In the press release `BlackLine Announces Third Quarter 2025 Financial Results', take the individual operating
expense line of the Condensed Consolidated Statements of Operations with the smallest amount in the nine months
ended September 30, 2025 (excluding the total line). By what percentage did it grow against the same period of
2024? Give it to one decimal place. & 6.1\% & 398.3\% \\
\addlinespace
In `Addus HomeCare Announces Third Quarter 2025 Financial Results', suppose adjusted EBITDA for the third quarter
of 2025 had grown year over year at the rate the Overview gives for net service revenues rather than at its own
rate. What would adjusted EBITDA for the quarter have been, to the nearest \$0.1 million? &
\$43.1 million & \$42.9 million \\
\addlinespace
In the Eurostat release `Industrial production down by 0.5\% in the euro area and by 0.3\% in the EU', take the
grouping with the smallest increase in the euro area bullet list for August 2026 compared with July 2026. In the
euro area block of the `compared with same month of the previous year' table, by how many percentage points is its
August 2026 figure above its March 2026 figure? &
0.4 percentage points & 5.5 percentage points \\
\addlinespace
In the Child Nutrition Programs: Income Eligibility Guidelines (July 1, 2025--June 30, 2026), for a household of
two, which of the chart's three area blocks gives the lowest Reduced Price Meals--185\% annual figure, and how much
lower than the next-lowest block's figure is it? &
Hawaii; \$4,096 lower & 48 Contiguous States, \$5,864 lower \\
\bottomrule
\end{tabular}
\caption{Representative held-out EDO questions that the base model answers incorrectly and the
hypernetwork-initialized adapters answer correctly. Question text is reproduced verbatim.}
\label{tab:examples}
\end{table*}

Table~\ref{tab:examples} illustrates the complexity of the target workload. The questions combine document
retrieval with conditional filtering, comparison, ranking, arithmetic, and cross-table lookup.

\subsection{Performance by number of relevant items}
\label{sec:hard}

Across documents, adaptation most often corrects structural errors over small sets: assigning a property to the
right entity, identifying members of a named set, or ordering values expressed in prose. Correction rates fall and
regression rates rise as questions involve more items.

\begin{table}[t]
\centering
\small
\begin{tabular}{lrrrrr}
\toprule
items & $n$ & base & microfit & corrected & regressed \\
\midrule
1--3   & 51 & .42 & .80 & .70 & .05 \\
4--6   & 61 & .43 & .64 & .57 & .27 \\
7--10  & 55 & .39 & .55 & .46 & .30 \\
11--15 & 31 & .32 & .42 & .33 & .40 \\
\bottomrule
\end{tabular}
\caption{Performance on the targeted evaluation questions, grouped by the number of items involved.
``Corrected'' denotes the proportion of base-model errors corrected by the adapter; ``regressed'' denotes the
proportion of correct base-model answers lost after adaptation.}
\label{tab:setsize}
\end{table}

In Table~\ref{tab:setsize}, base-model accuracy is fairly stable across the item-count groups, but correction rates
fall and regression rates rise as questions involve more items. On a separate slice
of 48 questions requiring a filter, sum, or count over an entire structure, accuracy remains near zero with and
without adaptation. The adapters therefore improve short structural operations but do not add the serial
computation needed for extended tallies.

\subsection{Interpretation}

The findings of \citet{allenzhu2023physics31,allenzhu2023physics32} distinguish the storage and retrieval of facts
from their manipulation. A model may retrieve individual facts successfully while failing to classify, compare,
or rank them without explicit intermediate steps. Our results suggest that document-specific training encodes
useful local relationships in the adapter---associations between entities and attributes, membership of named sets,
and orderings within short lists---reducing the manipulation required at inference. As the number of relevant items
grows, the model must carry out more intermediate operations, and the advantage declines. Microfitting therefore appears
to improve access to a document's local structure rather than extending the model's general capacity for serial
reasoning.

\section{Continual adaptation and future work}
\label{sec:compounding}

The current feedback loop revises the question generator but leaves the training objective and most
hyperparameters fixed. Future work will expand both:

\begin{itemize}
\setlength{\itemsep}{3pt}
\setlength{\parskip}{0pt}
\item \textbf{Corpus-wide initialization and meta-learning.} In deployments with sufficient data, a one-time
corpus-wide stage could adapt the general hypernetwork checkpoint to an organization's document and task
distribution before microfitting in small batches. This could raise the starting accuracy of generated adapters
and amortize domain adaptation across thousands of documents, allowing each small-batch run to focus on
document-specific gains. A first-order meta-learning experiment at development scale already improves initial
performance by a few percentage points, suggesting that prior documents can strengthen this initialization.

\item \textbf{Domain-adaptive training configuration.} The current system revises the question generator while
largely fixing learning rate, annealing, batch size, questions per document, and related hyperparameters. A
fine-grained sweep over questions per document, especially the untested range between forty and approximately
eighty-five, should determine where gains saturate and whether excess questions cause overfitting. Preliminary
experiments also find gains from mixing question-answer training with next-token prediction over documents and
their summaries. A broader controller could tune these choices alongside the training-objective mixture, reasoning
level, and context-window size for each model and data environment using small pilot runs and held-out adapter
performance.

\item \textbf{Document-conditioned generation.} Rather than assigning the same environment-level question mix to
every document, $G_t$ could allocate examples according to each document's structure, base-model errors, and
opportunities for improvement. A judge or learned utility model could rank candidate examples, while adaptive
allocation methods balance document-specific opportunities against the distribution of real user questions $U$.

\item \textbf{Drift and changing workloads.} The current EDO demonstration varies documents while
holding the target-question generator fixed. Longitudinal experiments will shift the workload examples $U$ and
measure how quickly $\Phi$ revises $r_t$, $G_t$ rebalances $T_{t,i}$, and adapter performance recovers.
\end{itemize}

\section{Limitations}
\label{sec:limitations}

\noindent\textbf{Document and domain scope.} Our evaluation is limited to documents of 2.2K--4K tokens; performance with
longer contexts remains to be evaluated. We observe the same broad pattern across multiple enterprise-style
document sets, higher-volume Gemma ablations, and a preliminary application of the setup to adapters in a retrieval
setting. However, performance on substantially different enterprise corpora and entirely new domains remains to be
established.

\noindent\textbf{Experimental scale.} The primary continual-improvement sequence contains three ten-document iterations,
one training seed per condition, and one base-model run per iteration. Its monotonic increase in relative gain is
based on three samples and does not by itself establish the contribution of each individual recipe revision.

\noindent\textbf{Attribution and controls.} The correctable-error study changes $V_{t,i}$ between recipes and should
therefore be interpreted as adaptive targeting rather than performance on a fixed workload. Control experiments show
that some gains arise from broader task adaptation. Future work should use systematic target-document,
unrelated-document, and shared-adapter controls to isolate and quantify the advantage of document-specific fitting.

\noindent\textbf{Workload drift.} We have not tested changes to workload examples $U$. The time required for $\Phi$ and $G_t$
to rebalance generated questions and adapt to a changing user-question distribution remains unmeasured.

\section{Conclusion}

Microfitting turns document-specific test-time training into a reusable offline process: it generates training
questions, fits and caches one adapter per document, and revises the generator from measured outcomes. Parameter
isolation lets the system add new documents without overwriting earlier adapters. Across fresh EDO batches,
successive recipe revisions produce stronger adapters. A separate study
shows that the same process can concentrate training on harder but still correctable model failures. Taken together,
these results suggest that organizations can continually specialize a frozen model to private document collections
using modest local hardware. Microfitting therefore offers a practical path to more accurate enterprise Q\&A
without repeated full-model training or datacenter-scale resources.

\section*{Contributions}

Nick Ryan led the research: he developed the microfitting method and feedback-driven data-generation pipeline, ran
the training runs, evaluations, and analyses, implemented the systems optimizations, and wrote the paper. Peter A.
Devine led the development of the Internalizer architecture on which this work is built and provided the pretrained
checkpoint. Peter A. Devine, Ben Sirb, and Hannah Sansford consulted on the work, providing feedback and research
direction, and reviewed the paper prior to its release. Alex Chiocchi supervised the work.

\bibliography{custom,new_refs,microfitting_refs}

\appendix

\section{Hypernetwork fine-tuning}
\label{app:hypernet}

The Internalizer \citep{devine2026internalizer} maps documents of up to 4K tokens to LoRA adapters for DeepSeek V4
Flash. The checkpoint used here was initially trained for document question answering with only the question
provided at inference; this describes the training setup, not a restriction on the intended applications.
We continued training on simple retrieval questions over synthetic enterprise documents to obtain a general
checkpoint for subsequent adaptation. Each document has five questions: four presented open-book and one
closed-book. Training used 2,000 steps of 64 examples, corresponding to approximately
25,600 documents, with the optimizer in Table~\ref{tab:hparams} and learning rates decayed to one-tenth of their
initial values over the final 500 steps. Open-book accuracy on reserved private documents increased during the
first few hundred steps and then stabilized, while closed-book accuracy declined as training shifted toward the
open-book setting. All microfitting experiments use the final checkpoint from this run.

\section{Output tokens per answer}
\label{app:latency}

On the 198 evaluation questions, the base model with thinking disabled produces a median of 3 output tokens per
answer, a mean of 5, and a ninetieth percentile of 11. With thinking enabled, the median is 378 tokens, the mean
is 657, the ninetieth percentile is 1,492, and the maximum is 7,026.

\end{document}